\documentclass[letterpaper, 10 pt, conference]{ieeeconf}  

\IEEEoverridecommandlockouts                              

\usepackage{siunitx}
\usepackage{cite}
\usepackage[none]{hyphenat}
\usepackage{todonotes}
\usepackage{perpage}

\usepackage[firstpage]{draftwatermark}
\SetWatermarkText{\begin{tabular}{l}
    \\
    \copyright 2024 IEEE.  Personal use of this material is permitted. 
    Permission from IEEE must be obtained for all other uses, in any current or future media, including reprinting/republishing this material \\
    for advertising or promotional purposes, creating 
    new collective works, for resale or redistribution to servers or lists, or reuse of any copyrighted component of this work in other works.\\
    \\
    \scriptsize {IEEE INTERNATIONAL CONFERENCE ON ROBOTICS AND AUTOMATION 2024. PREPRINT VERSION. ACCEPTED JANUARY, 2024.}
    \end{tabular}}
\SetWatermarkColor[gray]{0}
\SetWatermarkFontSize{0.22cm}
\SetWatermarkVerCenter{0.5cm}
\SetWatermarkAngle{0}
\SetWatermarkHorCenter{10.3cm}

\MakePerPage{footnote}

\title{\LARGE \bf \sc
RicMonk: A Three-Link Brachiation Robot with Passive Grippers for Energy-Efficient Brachiation
}

\author{Shourie S.  Grama$^{1*}$, Mahdi Javadi$^{1*}$, Shivesh Kumar$^{1}$, Hossein Zamani Boroujeni$^{1}$ and Frank Kirchner$^{1,2}$
\thanks{This work has been supported by the M-RoCK (FKZ01IW21002) and VeryHuman (FKZ01IW20004) projects funded by the German Aerospace Center (DLR) with federal funds from the Federal Ministry of Education and Research (BMBF) and is additionally supported with project funds from the federal state of Bremen for setting up the Underactuated Robotics Lab (201-342-04-2/2021-4-1).}
\thanks{$^{1}$Shourie S.  Grama, Mahdi Javadi, Shivesh Kumar, Hossein Zamani Boroujeni, Frank Kirchner are with Robotics Innovation Center, DFKI GmbH, 28359 Bremen, Germany 
(E-mail: \text{firstname.lastname@dfki.de}).
        \textit{Corresponding author: Shourie S. Grama}
}
\thanks{$^*$ The authors contributed equally to the work}
\thanks{$^{2}$Frank Kirchner is also with AG Robotik, University of Bremen, 28359 Bremen, Germany.}%
\thanks{The authors acknowledge the support of Dr. Melya Boukheddimi, Shubham Vyas, and Rohit Kumar.}
}

\begin{document}

\maketitle
\thispagestyle{empty}
\pagestyle{empty}

\begin{abstract}
    
    This paper presents the design, analysis, and performance evaluation of RicMonk, a novel three-link brachiation robot equipped with passive hook-shaped grippers. 
    Brachiation, an agile and energy-efficient mode of locomotion observed in primates, has inspired the development of RicMonk to explore versatile locomotion and maneuvers on ladder-like structures. 
    The robot's anatomical resemblance to gibbons and the integration of a tail mechanism for energy injection contribute to its unique capabilities. 
    The paper discusses the use of the Direct Collocation methodology for optimizing trajectories for the robot's dynamic behaviors and stabilization of these trajectories using a Time-varying Linear Quadratic Regulator. 
    With RicMonk we demonstrate bidirectional brachiation, and provide comparative analysis with its predecessor, AcroMonk - a two-link brachiation robot, to demonstrate that the presence of a passive tail helps improve energy efficiency. 
    The system design, controllers, and software implementation are publicly available on GitHub\footnote{The open-source implementation is available at \href{https://github.com/dfki-ric-underactuated-lab/ricmonk}{https://github.com/dfki-ric-underactuated-lab/ricmonk} and the video demonstration of the experiments can be viewed in accompanying video.}.
    
\end{abstract}
\section{\textsc{Introduction}}
\label{section:introuction}
    Brachiation, a mode common among primates like long-armed gibbons, involves swinging between tree branches using their arms. 
    Gibbons exhibit astonishing agility, showcasing their ability to brachiate at speeds reaching up to \SI{25}{\kilo\meter\per\hour}.  
    This locomotion mode holds potential for diverse applications such as agriculture surveillance, forest exploration, and biomimetic design. 
    Robots capable of brachiating and walking, known as Multi-locomotion robots (MLR), offer intriguing opportunities for study and implementation~\cite{fukuda_multi-locomotion_2005},~\cite{fukuda2012multi},~\cite{lu_dynamic_2010}. 
    Brachiation includes ``Slow Brachiation", where the robot swings between branches without a free-flight phase, and ``Fast Brachiation" or ``Ricochetal Brachiation" involving dynamic free-flight phases~\cite{nakanishi2000leaping}. 
    This unique locomotion bears similarity to walking and running, relying on full-body coordination for balance and propulsion. 
    The adoption of underactuated control for brachiation exploits their dynamics despite challenges posed by nonlinear control.

    \begin{figure}
        \centering
        \includegraphics[width=\linewidth]{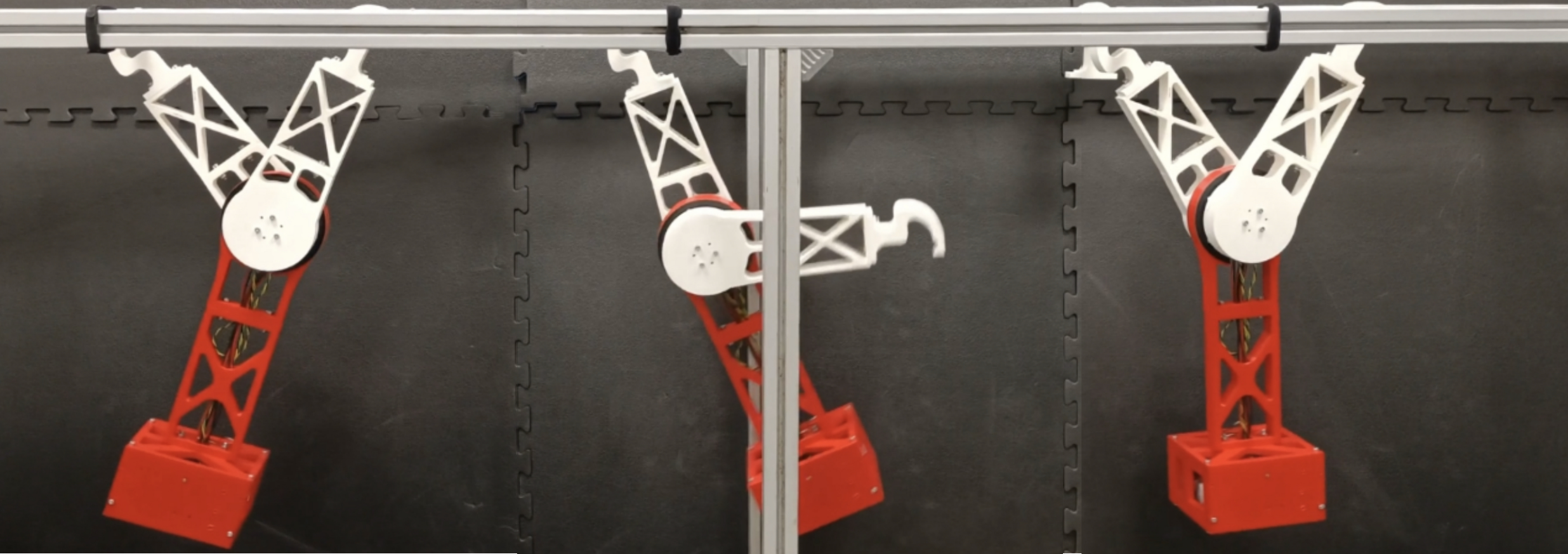}
        \caption{RicMonk performing brachiation maneuvers}
        \label{fig:BF}
        \vspace{-0.2cm}
    \end{figure}

    Research in brachiating mechanisms has a longstanding history. 
    An early study involved simulating and analyzing a single cycle of motion for a five degrees of freedom~(DOF) brachiation robot~\cite{fukuda_brachiation_1991}. 
    Subsequent studies concentrated on simpler designs, typically comprising two or three links, and offered experimental validation for the proposed algorithms~\cite{fukuda-target-dynamics},~\cite{yang_design_2019},~\cite{meghdari2013minimum}. 
    These designs exhibited favorable power-to-weight ratios, enabling efficient and low mechanical cost momentum generation~\cite{shata2019brachiating}.     
    Early studies introduced acrobot-type underactuated robots equipped with actuated grippers, emphasizing heuristic methods for behavior generation~\cite{yamafuji_study_1992}, \cite{saito_swing_1993}, \cite{saito_heuristic_1994}. 
    The motion along a flexible rope offers energy-efficient advantages by including the cable dynamics in the robot's model as a compound double pendlum~\cite{farzan_feedback_2019}. 
    Experimental validation by taking advantage of Time-Varying Linear Quadratic Regulator (TVLQR) and Sum-of-Squares (SOS) optimization highlights the importance of the energy optimal model-based controllers~\cite{farzan2020robust}. 
    Similarly, a three-link robot design focused on discontinuous brachiation motion, optimizing design parameters for energy efficiency is described in~\cite{yang_design_2019}. 
    The role of the tail in exciting the robot's natural frequency is a crucial factor in achieving ricochetal transverse brachiation on the real system~\cite{lin_design_2017}. 
    A comprehensive overview of the previous research on brachiation-type mobile robots from an energy point of view is presented in~\cite{andreuchetti_survey_2021}. 
    The common feature among all these brachiation robots is the presence of active grippers, which introduces additional complexities to the system design, and maintenance and energy expenditure. 
    AcroMonk is the first underactuated two-link robot capable of energy-optimal continuous brachiation with passive grippers over a horizontally-laid ladder bar~\cite{javadi2023acromonk}. 
    Even though the robot is capable of an unlimited number of brachiation maneuvers in experiments, it is only able to perform robust forward brachiation. 
    This paper introduces RicMonk, a three-link underactuated robot with passive grippers capable of energy-optimal bidirectional brachiation as shown in~Fig.~\ref{fig:BF}. 
    Finally, we present a comparative analysis with AcroMonk by emphasizing the tail's role in enhancing energy-efficient brachiation using TVLQR, and the gripper improvement. 
    In the spirit of~\cite{javadi2023acromonk},~\cite{2023_ram_wiebe_double_pendulum},~\cite{2022_rss_realaigym}, and~\cite{Wiebe2022}, the project is open-sourced\footnote{
        \href{https://github.com/dfki-ric-underactuated-lab/ricmonk}{https://github.com/dfki-ric-underactuated-lab/ricmonk}
    } to help researchers and provide them with a low-cost kit to encourage the study of underactuated robotics. 
    The hardware test demonstrations are available and can be viewed in the accompanying video.

    \paragraph*{Organization}    
    Section \ref{section:mechatronics} addresses the mechatronics design of the RicMonk. 
    Trajectory generation for energy-optimal brachiation is presented in Section \ref{section:trajectoryOptmization}. 
    Section \ref{section:trajectoryStabilization} details the stabilization methods utilized for robot control. 
    Results are presented in section \ref{section:experimentalValidation} and finally, a conclusion is provided in Section \ref{section:conclusion}.

\section{\textsc{Mechatronics}}
\label{section:mechatronics}
The process of Mechatronic integration for a robotic system involves incorporating mechanical design and essential electrical components to ensure its operational functionality, capability, and safety. 
As envisioned in the preceding generation of the RicMonk, the AcroMonk~\cite{javadi2023acromonk}, the design is intended to remain compact and wireless and needs to be made of readily available and low-cost materials. 
Utilizing a single actuator in AcroMonk was a deliberate decision, but it resulted in a trade-off. 
The price paid for this simplicity is evident in its inability to perform multiple backward brachiation maneuvers robustly, as the arms could not be commanded independently to release the bar without the risk of falling. 
Furthermore, AcroMonk cannot dynamically to build up momentum like a monkey. 
The design of RicMonk takes these shortcomings into account and will be discussed in detail in the following sections. 
To address these shortcomings, RicMonk is equipped with two actuators and a tail that mimics a monkey's body, as depicted in Fig.~\ref{fig:BF}. 
The subsequent sections provide detailed explanations of both mechanical and electrical system designs. 

                    

\begin{figure}
    \setlength{\abovecaptionskip}{0pt} 
    \setlength{\belowcaptionskip}{0pt} 
    \centering
    \includegraphics[width=0.7\linewidth]{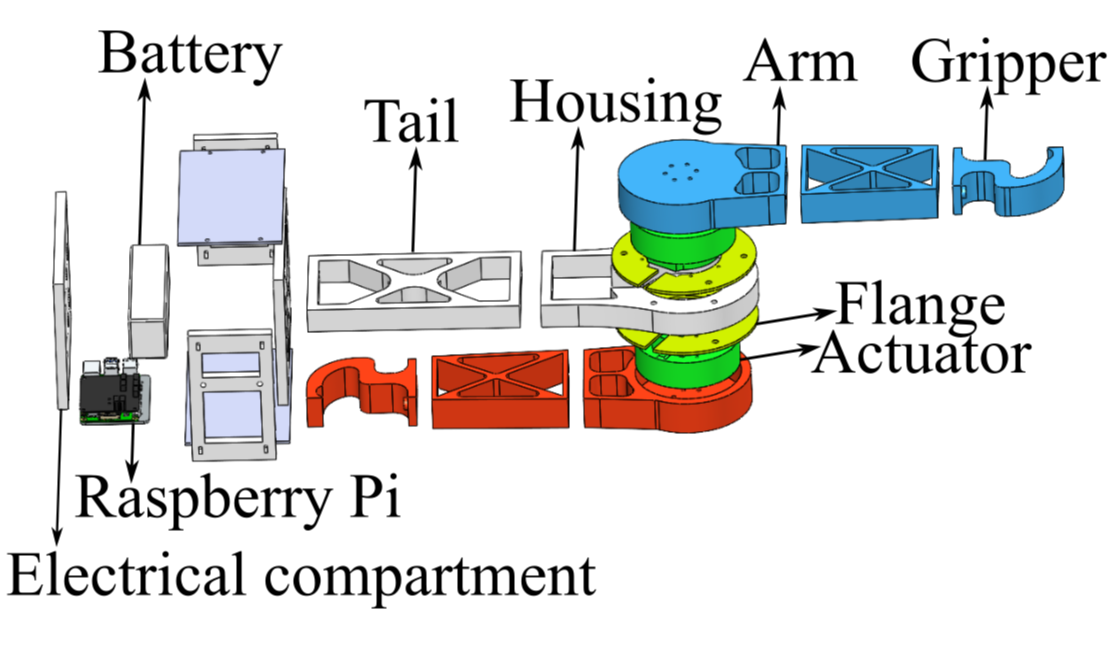}
    \caption{Exploded view of the RicMonk}
    \label{fig:RicMonkAndItsExplodedView}
\end{figure}

\subsection{Design strategy}
    Incorporating a compact design, RicMonk meets the demands for portability and agility, while its lightweight 3D-printed modular components allow for easy adjustment of the robot's inertial parameters. 
    RicMonk adopted the arm and gripper design from AcroMonk. 
    The tail was carefully designed to reduce wiring and tangling issues while providing a symmetrical housing for the integration of electrical and computing components. 
    Moreover, it acts as the point of connection for the housings of both motors, while each arm is directly linked to its corresponding motor shaft. 
    The exploded view of RicMonk along with the labeling of the components is presented in Fig.~\ref{fig:RicMonkAndItsExplodedView}. 
    An efficient gripper design is crucial in providing a stable passive pivotal joint as the robot swings. 
    AcroMonk consists of a passive gripper with a cylindrical groove of radius \SI{11.5}{\mm}. 
    Employing such a gripper for RicMonk, which weighs nearly twice as much as AcroMonk (due to the extra actuator and tail), induces instability, vibrations, and undesired motion during brachiation. 
    To resolve this, the cylindrical groove was replaced by a conical one, which has a larger radius of \SI{13}{\mm} and a smaller radius of \SI{11.5}{\mm}. 
    These values are obtained by considering the center of mass offset from the hooking surface. 
    This enhanced stability, and also increased torque requirement for releasing the arm from the support bar. 

\subsection{Electrical implementation}
    RicMonk is a three DOF underactuated brachiation robot equipped with two Quasi-Direct-Drive motors~\cite{mjbots} that allow independent arm movement. 
    With a 6:1 gear ratio, these quasi-direct drives soffer the advantage of delivering high torque and being back-drivable with low friction. 
    The motor controller boards enable communication with the actuators via Pi3hat~\cite{mjbots}, ensuring reliable communication and incorporating an onboard Inertia Measurement Unit~(IMU) (shown in Fig.~\ref{fig:RicMonkAndItsExplodedView}). 
    State estimation utilizes IMU measurements and actuator data to determine the state of the passive DOF with respect to the vertical plane, denoted as $q_1$ in Fig.~\ref{fig:fixedBase}. 
    A 3000 mAh Lithium polymer~(LiPo) battery is integrated as the power source to satisfy the portability design demand. 
    For safety purpose, a Radio Communication Transmitter Protocol~(TX protocol) is employed to incorporate a relay for turning off motors during erratic behavior. 

\section{\textsc{Behavior State Machine}}
\label{section:trajectoryOptmization}
\begin{figure}[tb]
    \centering
    \includegraphics[scale=0.25]{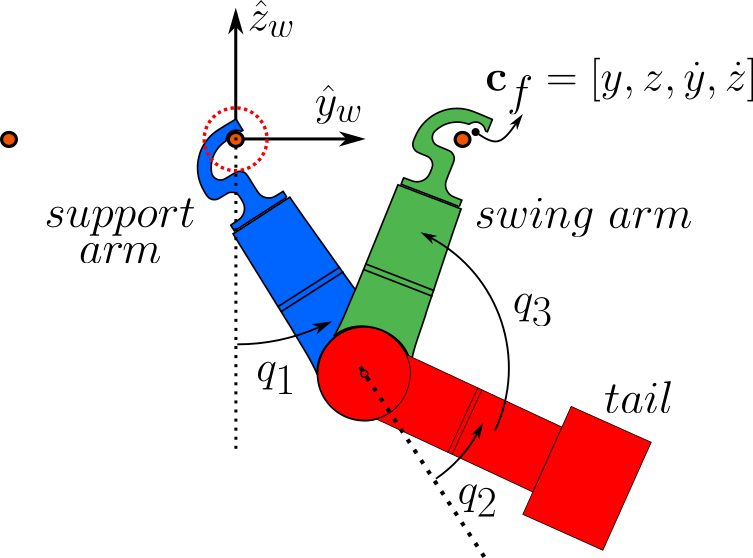}
    \caption{Fixed base model of the robot}
    \label{fig:fixedBase}
\end{figure}

Javadi \textit{et al.}~\cite{javadi2023acromonk} proposed a state machine to ensure robust and continuous forward or backward brachiation while accounting for potential failures. 
RicMonk also employs a similar state machine that integrates \textit{gripper-related heuristics} with a swing phase to generate a maneuver. 
The swing motion is associated with a sequence of transitions among three fixed configurations, referred to as \textit{atomic behaviors}. 
These points are determined by hanging or double support configuration of the robot known as Zero-to-Front~(ZF), Zero-to-Back~(ZB), Front-to-Back~(FB), and Back-to-Front~(BF). 
This approach enables uninterrupted robust brachiation even amidst disruptions and also simplifies trajectory optimization. 
For swing behavior generation, a fixed-base robot model is used as illustrated in Fig.~\ref{fig:fixedBase}. 

\begin{figure}[tb]
    \centering
    \includegraphics[scale=0.2]{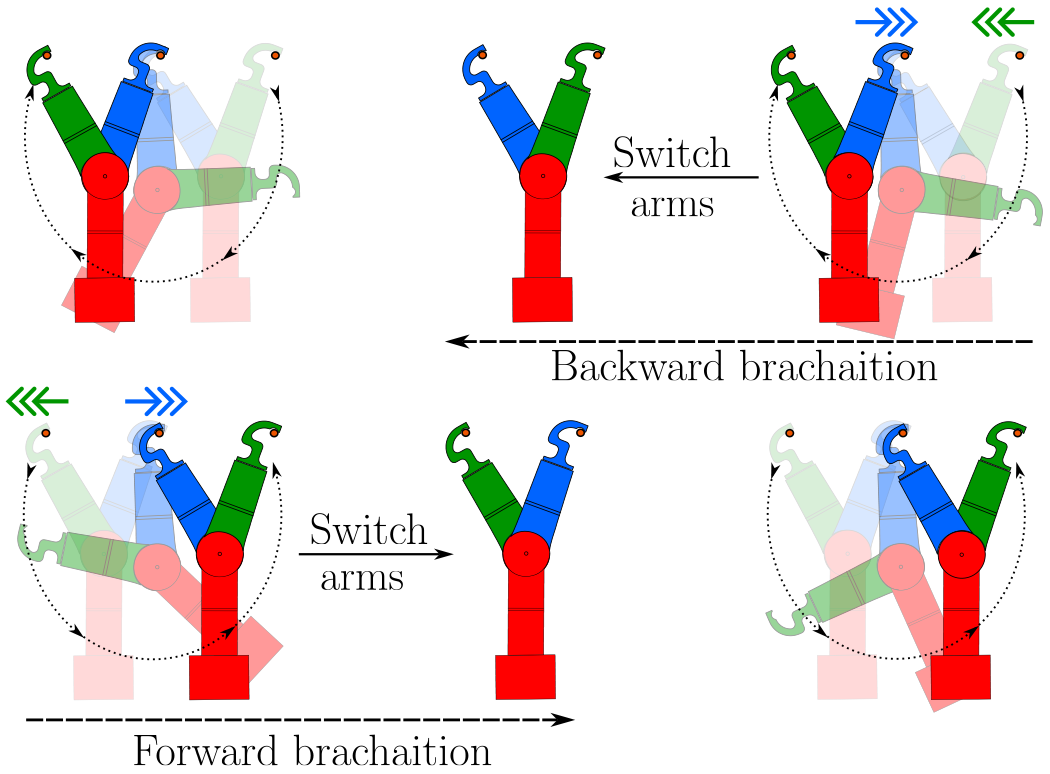}
    \caption{Illustration of bi-directional brachiation maneuver}
    \label{fig:bidirectionalBrachiation}
\end{figure}

In this model for RicMonk, a virtual joint connecting the support arm to the ladder bar at $\mathbf{(y_{c}, z_{c})}$ is depicted as a revolute joint (indicated by the dotted circle in Fig. \ref{fig:fixedBase}). 
The arms are distinguished by different colors, with blue denoting the support arm and green for the swing arm. 
The support arm holds the bar and forms a passive revolute joint during maneuvers, while the swing arm can move freely to catch the target bar. 
Gripper-related heuristics involve the actions of releasing and catching the front and back bars, which are abbreviated as Front Release (FR), Back Release (BR), Front Catch (FC), and Back Catch (BC). 
A complete motion cycle involves a coordinated sequence of actions performed by the swing arm, including the release, swing maneuver, and catch phases. 
Combining these sequences and alternating between support and swing arms in the cyclic motion results in bi-directional brachiation as illustrated in Fig.~\ref{fig:bidirectionalBrachiation}. 
For example, executing FR~$\rightarrow$~FB~$\rightarrow$~BC~$\rightarrow$~switch~arms~$\rightarrow$~FR~$\rightarrow$~FB~$\rightarrow$~BC results in two cycles of backward brachiation. 
The following subsections provide a detailed explanation of swing behavior generation and highlight the heuristics with two actuators for RicMonk.

\subsection{Gripper Heuristics}
To catch and release from the bars with passive grippers, it is necessary to apply torques to the arms in appropriate directions. 
The colored arrows in Fig.~\ref{fig:bidirectionalBrachiation} illustrates the force directions for front and back release of the swing and support arm. 
For the back release, the support arm torque (\SI{4.3}{\newton\meter}) is higher than the swing arm torque (\SI{2.3}{\newton\meter}), whereas for the front release, the situation is reversed, with the support arm torque (\SI{2.2}{\newton\meter}) being lower than the swing arm torque (\SI{6.3}{\newton\meter}). 

The swing arm needs to grab the bar from above for FC, while the gripper needs to grab it from below for BC. 
To obtain the initial state for the trajectory optimization of these maneuvers, iterative release experiments are conducted and the robot states are collected. 
The mean value for BR is $\mathbf{x}_0^{\text{BR}} = [-0.59, 0.68, -2.39, -0.37, -0.32, 1.76]$ with $\boldsymbol{\sigma}^{\text{BR}}_{0} = [0.02, 0.03, 0.05, 0.24, 0.27, 0.6]$. 
Similarly, for FR the mean value is $\mathbf{x}_0^{\text{FR}} = [0.57, -0.57, 2.76, 0.68, -1.47, 4.24]$ with  $\boldsymbol{\sigma}^{\text{FR}}_{0} = [0.01, 0.03, 0.03, 0.42, 0.29, 1.8]$. 
Statistical analysis suggests that $\mathbf{x}_0^{\text{BR}}$ and $\mathbf{x}_0^{\text{FR}}$ are sufficiently reliable initial states for trajectory optimization of BF and FB respectively. 

\subsection{Trajectory Optimization for Swing Behavior}
The presence of an onboard battery underscores the importance of energy-efficient brachiation while respecting dynamics, actuation, and task space limitations. 
This can be formulated as an optimal control problem, involving the minimization of an objective function within a finite time frame while respecting specified constraints. 
Trajectory optimization for the swing behavior spans the period between the release and catch phases. 
It begins right after the release action and continues until just before the swing arm grabs the bar. 
The state of the robot, $\mathbf{x}=[\mathbf{q}, \dot{\mathbf{q}}]$, consists of the generalized position vector ($\mathbf{q}=[q_1, q_2, q_3] \in \mathbb{R}^3$) and its first order time derivative. 
The input vector of the robot, $\mathbf{u}=[u_1, u_2] \in \mathbb{R}^2$, consists of the torque input values for the actuated joints that control $q_1 + q_2$ and $q_3$. 
The dynamics of the system are given by Eq.~\ref{eq:dynamics}, where $\mathbf{M}(\mathbf{q})$ is the mass-inertia matrix of the robot, $\mathbf{C}(\mathbf{q}, \dot{\mathbf{q}})$ is the matrix that describes the Coriolis forces, $\tau_g(\mathbf{q})$ is the gravity vector, and $\mathbf{B}$ is the actuator selection matrix.
\begin{equation}
    \mathbf{M}(\mathbf{q})\ddot{\mathbf{q}} + \mathbf{C}(\mathbf{q}, \dot{\mathbf{q}})\dot{\mathbf{q}} = \tau_g(\mathbf{q}) + \mathbf{B}\mathbf{u}
    \label{eq:dynamics}
\end{equation}

\noindent The formulation of the mathematical problem for trajectory optimization is presented in Eq.~\ref{eq:trajopt}. 
The cost function contains state and input regularization terms, where the weighting matrices $\mathbf{K}$ and $\mathbf{R}$  penalize high joint velocities and input torques. 
Eq.~\ref{eq:dynamicConstraint} is the set of first-order ordinary differential equations that captures the system dynamics as shown in Eq.~\ref{eq:dynamics}. 
\begin{subequations}
    \begin{equation}
        \label{eq:trajopt}
        \begin{aligned}
            &\min_{\mathbf{x}, \mathbf{u}} \int_{0}^{T} (\dot{\mathbf{q}}^{T}\mathbf{K}\dot{\mathbf{q}} + \mathbf{u}^{T}\mathbf{R}\mathbf{u})\ dt\\
            &\mathrm{subject\ to}:
        \end{aligned}
    \end{equation}
        \begin{align}
            & \dot{\mathbf{x}} = \mathbf{f}(\mathbf{x}, \mathbf{u}) \label{eq:dynamicConstraint}\\
            & g(\mathbf{x}, \mathbf{u}, t) \leq 0 \label{eq:inequalityConstraint}\\
            & \mathbf{x}(t_0) = \mathbf{x}_0 \label{eq:initialState}\\
            & \mathbf{x}(t_f) \text{ defined using } \mathbf{c}_f^*\label{eq:finalState} 
        \end{align}
\end{subequations}
All the inequality constraints like position, velocity or torque limits, and collision constraints are represented by Eq.~\ref{eq:inequalityConstraint}. 
The initial state ($\mathbf{x}_0$) of the robot for the ZF and ZB maneuver is zero and the robot builds up energy to reach the final state. 
For BF and FB maneuvers, the initial state of the robot is obtained based on state information from iterative release experiments with the robot. 
As previously mentioned, both FB and BF swings commence after a release behavior, which means the initial state is not static as in ZB and ZF. 
As mentioned in Section~\ref{section:mechatronics}, the connection of the tail to arms results in a non-serial kinematic coupling. 
Due to this, the inverse geometry cannot directly provide the robot's final state $\mathbf{x}(t_f)$, which is necessary for the trajectory optimization as stated in Eq.~\ref{eq:finalState}. 
The methodology to obtain the $\mathbf{x}(t_f)$ includes usage of the desired final cartesian position and velocity of the gripper of the swing arm, i.e. $\mathbf{c}_f^*=[y^*, z^*, \dot{y}^*, \dot{z}^*]$. 
At each iteration, the solver calculates the forward kinematics to obtain the task space coordinates of the swing arm $\mathbf{c}_f$ as shown in Fig.~\ref{fig:fixedBase}, based on the state's decision variable and tries to minimize the error compared to $\mathbf{c}_f^*$. 
Taking advantage of the Direct Collocation method by \textsc{Drake}~\cite{drake} toolbox, the trajectories are optimized for the swing phases using the Sparse Nonlinear OPTimizer~(SNOPT)~\cite{gill2005snopt}. 
Hyperparameters for each of the maneuvers are detailed in Table \ref{table:hyperparameters}, with $N$ representing the number of collocation points and $\hat{t}_f$ initial guess period in seconds. 
For optimal tracking of the trajectory, it is important to have smooth trajectories with minimal jerk, acceleration, and torque consumption. 
Considering the system's behavior, it is observed that the robot is unable to efficiently track trajectories when absolute jerk and acceleration exceed \SI[per-mode=symbol]{3000}{\radian\per\second\cubed} and \SI[per-mode=symbol]{200}{\radian\per\square\second} respectively. 
Trajectory profiles were carefully analyzed to ensure the compatibility of the obtained trajectories with the real system. 
In the initial iterations, the optimization problem is warm-started using a straight-line trajectory that connects the initial and final states with a first-order hold over an interval extending from zero to an estimated final time ($\hat{t}_f$ in seconds). 
After several iterations and utilizing agreeably smooth previously generated trajectories to warm-start the optimization problem, optimal trajectories were obtained for each of the maneuvers. 
The hyperparameters are carefully chosen to prevent impulsive peaks in the torque or joint velocity profiles. 
\begin{table*}[t]
    \scriptsize
    \centering
    \setlength{\tabcolsep}{5 pt} 
    \renewcommand{\arraystretch}{1.2} 
    \caption{Hyperparamters for trajectory optimization problem}
    \label{table:hyperparameters}

    \begin{tabular}{lcccc}
        \toprule
        Parameter & ZB & ZF & BF & FB \\
        
        \midrule
        N & 40 & 60 & 40 & 35\\            
        $\hat{t}_f$ & [0, 3.0] & [0, 3.2] & [0, 4] & [0, 1.5]\\
        $\mathbf{c}_f^*$ & [-0.345, 0.0, 0.83, 0.75] & [0.35, 0.025, 0.74, -1.4] &  [0.36, 0.03, 0.9, -1.2] & [-0.35, 0.0, 0.83, 0.75]\\
        $\mathbf{R}$ & diag[100, 7000] & diag[520, 440] & diag[6000, 5000] & diag[120, 120]\\
        $\mathbf{K}$ & diag[0, 100, 100] & diag[0, 200, 240] & diag[0, 500, 200] & diag[0, 94.5, 108]\\
        \bottomrule
    \end{tabular}
\end{table*}

The BF and FB trajectories generated could not be tracked by the controller, given they have an initial velocity and lower torque usage. 
The inability to successfully track trajectory points to the sim-to-reality gap. 
Several factors contribute to widening this gap, of which motor parameters like friction, damping, and rotor inertia play a major role. 
Even with perfect identification of these parameters, current solvers tend to disregard them when solving the optimization problem. 
Consequently, there is always a need for some heuristics to fill the sim-to-reality gap based on the behavior of the system. 
We gained inspiration by observing the behavior of brachiation masters like monkeys and gibbons. 
These animals consistently aim above the bar, allowing for a brief moment of descent due to gravity. 
Similarly, when approaching a handhold from below, they first make contact with the surface and then grasp the bar from beneath.     
A similar heuristic approach is applied to enable RicMonk to successfully perform brachiation. 
It assumes a final configuration above the bar to facilitate falling over it, which also holds when approaching the bar from below. 

\section{\textsc{Trajectory Stabilization}}
\label{section:trajectoryStabilization}
    Trajectory stabilization is crucial for all robots to ensure trajectory tracking and it has been implemented in RicMonk to help the robot perform brachiation motion. 
    This section describes the model-based TVLQR that is employed to stabilize the optimal trajectories for the RicMonk.

    The controller linearizes the nonlinear dynamics shown in Eq. \ref{eq:dynamicConstraint} using first-order Taylor expansion in error coordinates that result in the state space vector system (Eq.~\ref{eq:linearizedDynamics}). 
    \begin{equation}
        \begin{gathered}
            \label{eq:linearizedDynamics}
            \dot{\Bar{\mathbf{x}}}(t) \approx \mathbf{A}(t)\Bar{\mathbf{x}}(t) + \mathbf{B}(t)\Bar{\mathbf{u}}(t)
            \\
            \text{with $\Bar{\mathbf{x}}=\mathbf{x}-\mathbf{x_0}$, $\Bar{\mathbf{u}}=\mathbf{u}-\mathbf{u_0}
            $}
        \end{gathered}
    \end{equation}
    TVLQR minimizes a time-varying cost function, $J(t)$ (Eq.~\ref{eq:TVLQRcost}), as it optimizes a varying control input, $\mathbf{u}(t)$. 
    The matrices $\mathbf{Q}(t)$, $\mathbf{R}(t)$, and $\mathbf{Q}_f(t)$ serve as weighting matrices crucial for stabilizing the trajectories. 
    Specifically, $\mathbf{Q}(t)$ and $\mathbf{R}(t)$ are responsible for weighting the state and control input, respectively, while $\mathbf{Q}_f(t)$ is the weighting matrix for the final state. 
    The optimal time-varying feedback matrix $\mathbf{K}(t)$ is obtained by solving the time-varying Riccati equation~\cite{underactuated}.       

     \begin{equation}
        \label{eq:TVLQRcost}
        J = \mathbf{\bar{x}}^T(t) \mathbf{Q}_f \mathbf{\bar{x}}(t) + \int_{0}^{t_f} \left( \mathbf{\bar{x}}^T(t) \mathbf{Q} \mathbf{\bar{x}}(t) + \mathbf{\bar{u}}^T(t) \mathbf{R} \mathbf{\bar{u}}(t)\right) \,\, dt
    \end{equation}   
    \begin{equation}
        \label{eq:TVLQRcontrolLaw}
        \mathbf{u}(t) = \mathbf{u}(t) -\mathbf{K}(t)\Bar{\mathbf{x}}(t)
    \end{equation}
    Mismatch between the mathematical model and the robot may affect the tracking performance for the TVLQR on the real system. 
    Approximation of highly non-linear systems could also add challenges to the controls. 
    However, TVLQR has the capability for real-time error compensation using previously obtained optimal control gains. 
    The TVLQR hyperparamters $\mathbf{Q}$ and $\mathbf{R}$ for various maneuvers are summarized in Table \ref{table:TVLQRParams} and $\mathbf{Q}_f=\mathbf{I}_{6\times6}$ for all maneuvers. 

    \begin{table}[t]
        \centering
        \caption{TVLQR hypermeters for the atomic behaviours}
        \label{table:TVLQRParams}
        \begin{tabular}{lcc}
        \toprule
        Maneuver & $\mathbf{Q}$ & $\mathbf{R}$ \\
        \midrule
        ZB  & diag[1, 1, 15, 1, 2, 1] & diag[8, 8]\\
        ZF  & diag[1, 1, 15, 1, 2, 1] & diag[5, 5]\\
        BF  & diag[1, 1, 17, 1, 2, 0.2] & diag[20, 20]\\
        FB  & diag[1, 1, 15, 1, 2, 1] & diag[20, 20]\\
        \bottomrule             
        \end{tabular}
    \end{table}

    TVLQR performed well with the ZB and ZF trajectory. However, to enable RicMonk to perform BF and FB maneuvers, heuristics were considered in trajectory optimization as mentioned in the Section~\ref{section:trajectoryOptmization} using TVLQR.

\section{\textsc{Experimental Validation}}
\label{section:experimentalValidation}
\begin{figure*}
        \centering
        \includegraphics[scale=0.5]{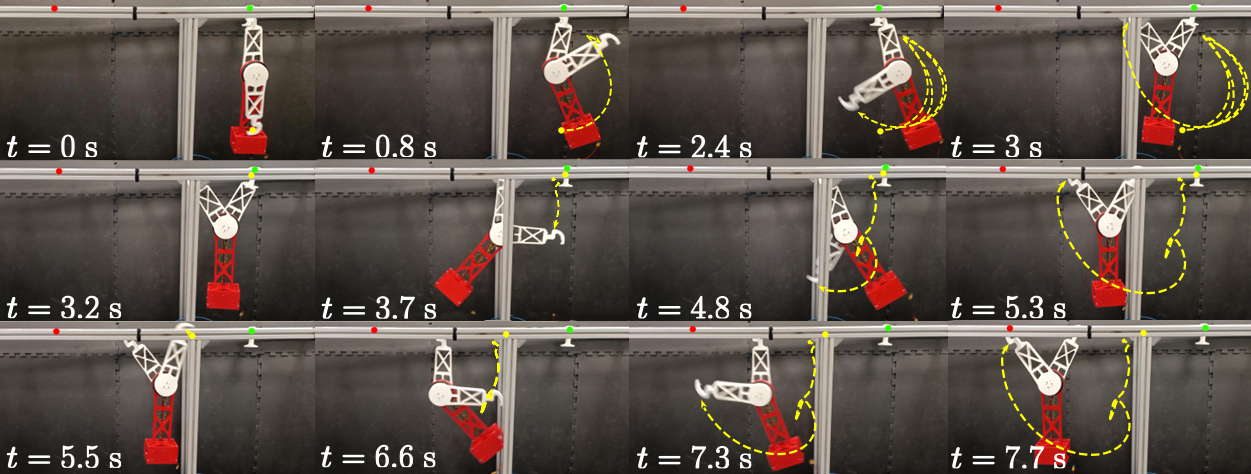}
        \caption{Snapshots of RicMonk performing multiple backward brachiation cycles. 
        First row corresponds to a ZB maneuver (left to right), while second and third rows correspond to consecutive FB maneuvers (left to right). 
        Green and red dots indicate start and end points respectively}
        \label{fig:threeBacBrach}
    \end{figure*}

\begin{figure*}
        \setlength{\abovecaptionskip}{0pt} 
        \setlength{\belowcaptionskip}{0pt} 
        \centering
        \begin{subfigure}[b]{0.32\textwidth}
            \centering
            \includegraphics[scale=0.2]{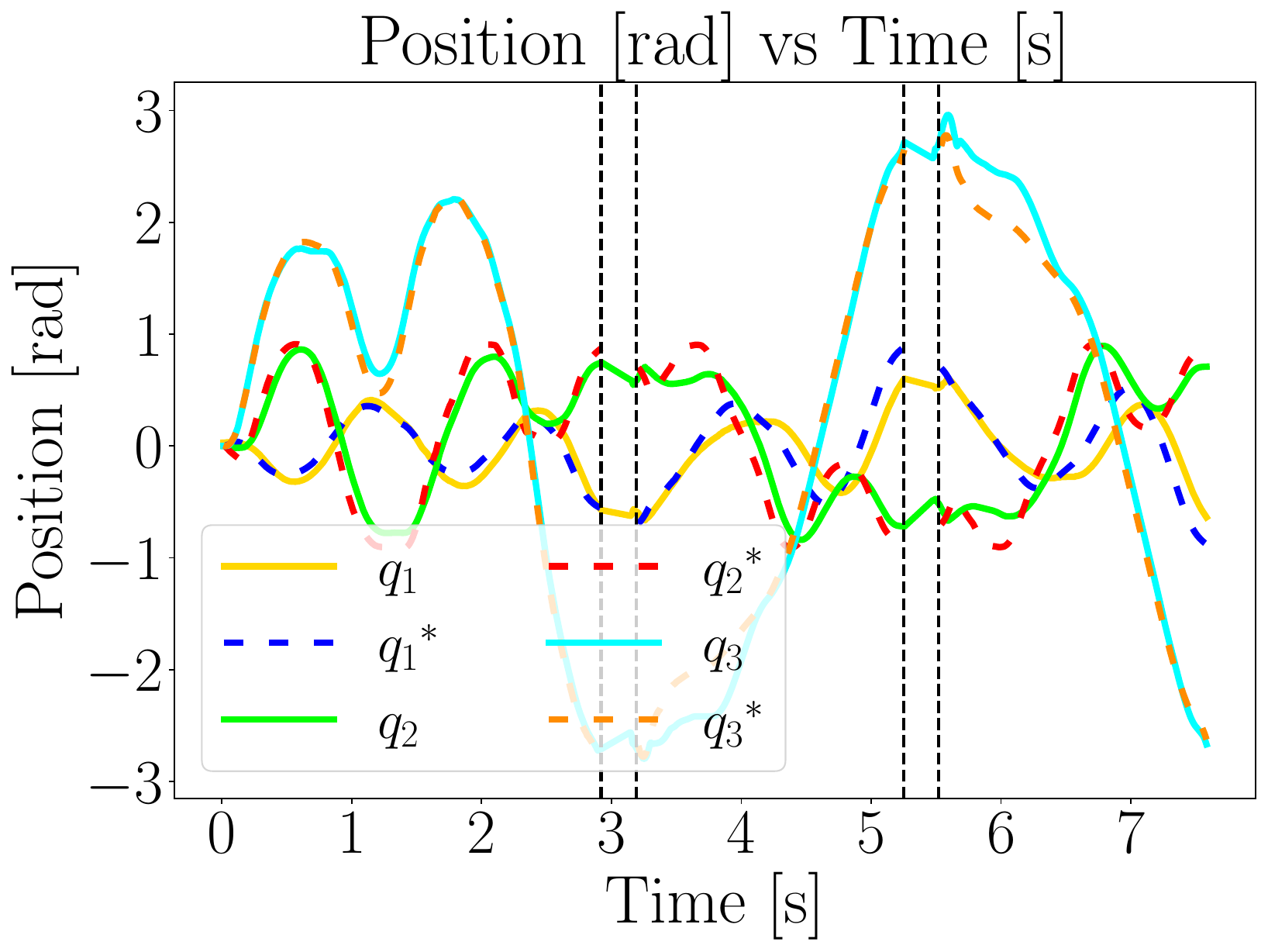}
            \caption{Generalized position plot}
            \label{fig:threeBacBrach_pos}
        \end{subfigure}
        \hfill
        \begin{subfigure}[b]{0.32\textwidth}
            \centering
            \includegraphics[scale=0.2]{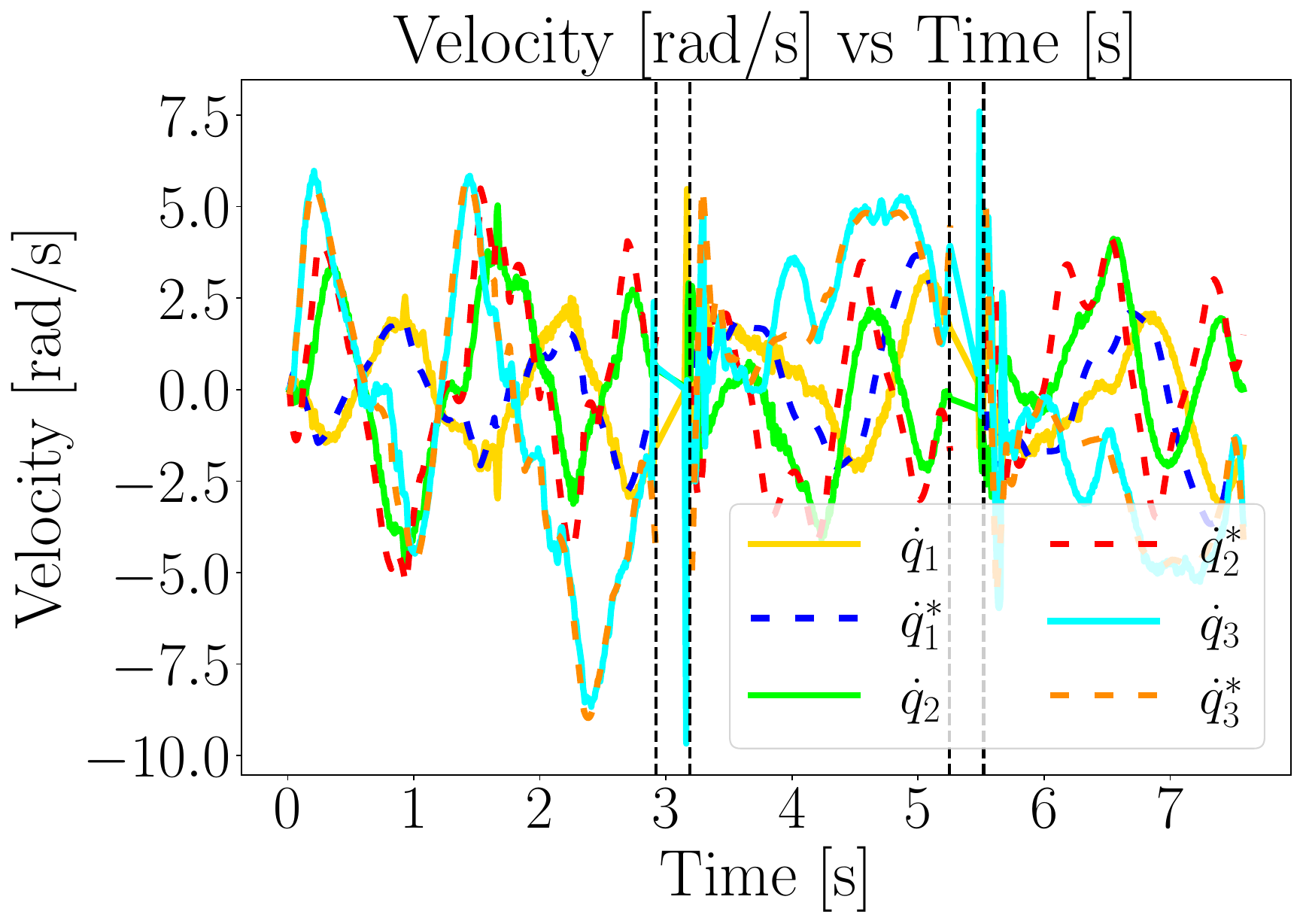}
            \caption{Generalized velocity plot}
            \label{fig:threeBacBrach_vel}
        \end{subfigure}
        \hfill        
        \begin{subfigure}[b]{0.32\textwidth}
            \centering
            \includegraphics[scale=0.2]{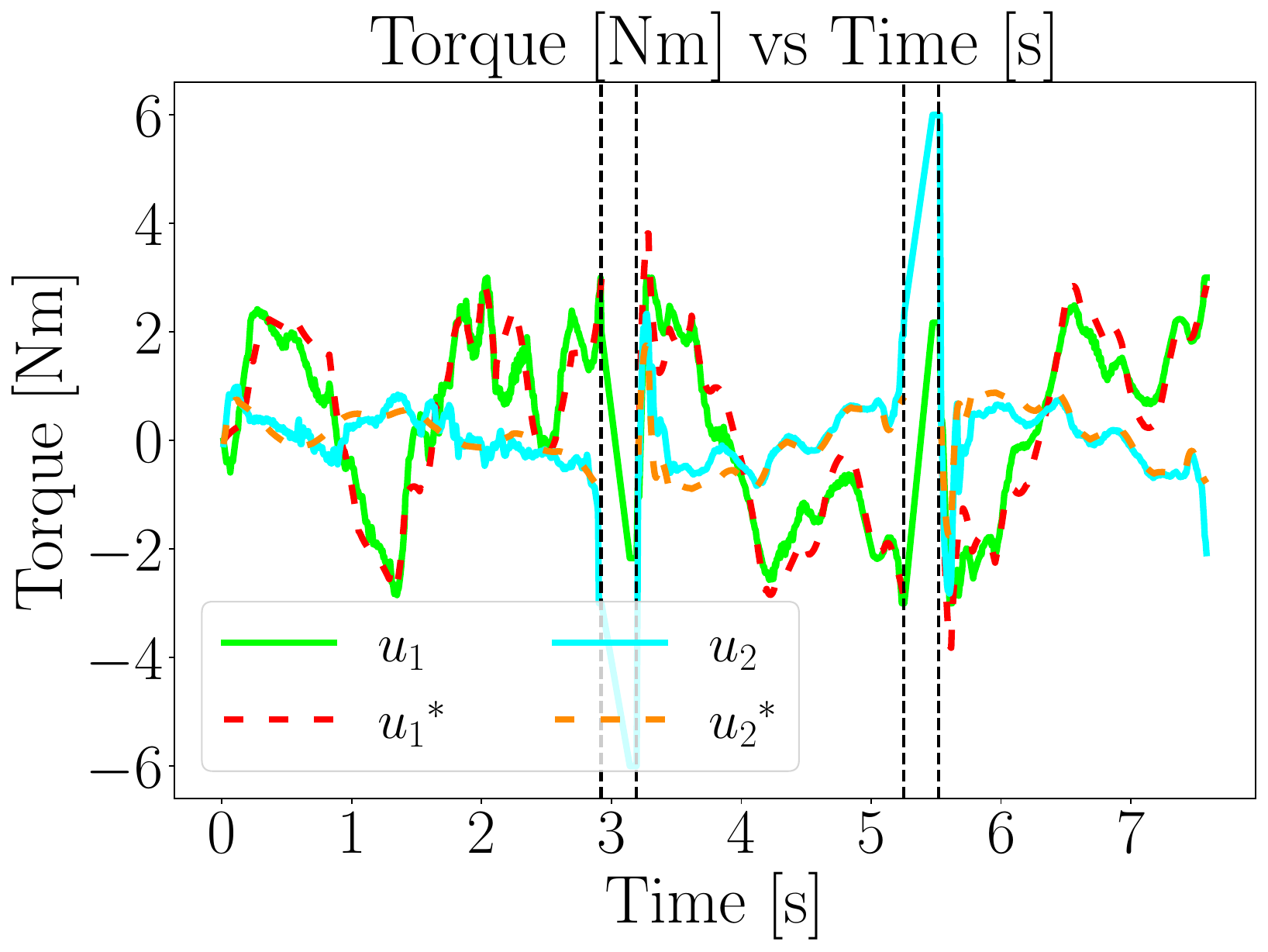}
            \caption{Optimum input plot}
            \label{fig:threeBacBrach_tau}
        \end{subfigure}
        
        \caption{Plots for the multiple backward brachiation maneuvers - one ZB maneuver, and two FB maneuvers in succession. 
        Vertical dotted lines separate swing phases and grasp and releases}
        \label{fig:plotThreeBacBrach}
    \end{figure*}

    All the atomic behaviors were realized on the real system with trajectory stabilization provided by the TVLQR. 
    These basic maneuvers were stitched together to enable the robot to perform multiple brachiation motions. 
    In Fig.~\ref{fig:threeBacBrach} RicMonk performs three consecutive backward brachiation maneuvers, namely, a ZB maneuver followed by two FB maneuvers. 
    Additionally, Fig.~\ref{fig:plotThreeBacBrach} provides plots representing the desired (indicated with superscript *) and measured positions, velocities, and input torques. 
    The plots demonstrate reasonable tracking of desired trajectories resulting in successful brachiation.  
        
    \subsection{Robustness Tests}

\begin{figure*}
            \centering
            \includegraphics[scale=0.514]{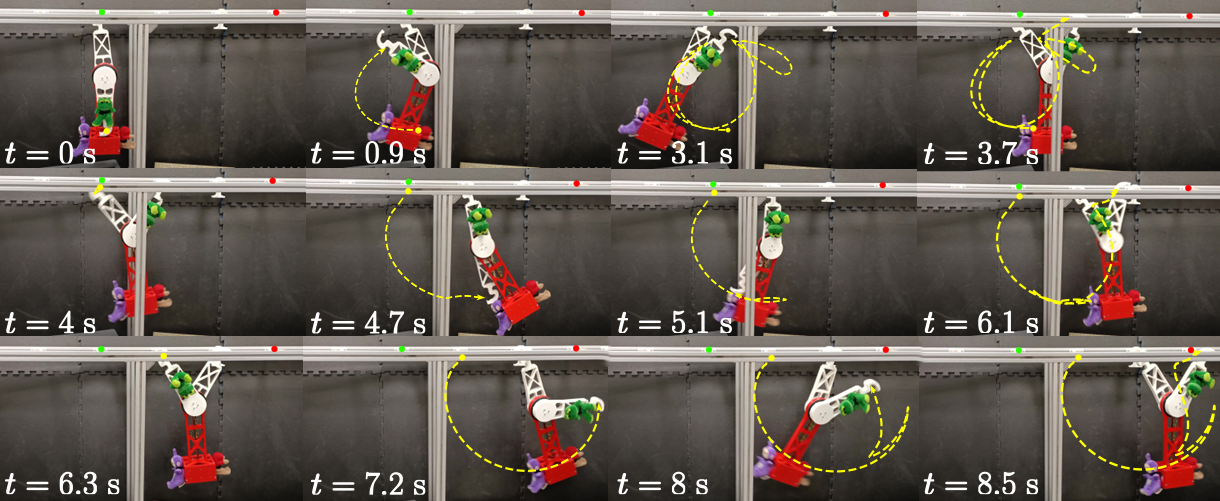}
            \caption{Snapshots of RicMonk performing multiple brachiation motions in the presence of mass uncertainty that is loaded asymmetrically. First row corresponds to a ZF maneuver (left to right), while second and third rows correspond to consecutive BF maneuvers (left to right). In row 2, second and third images illustrate the robot arm stuck due to a disturbance in motion. Green and red dots indicate start and end points respectively}
            \label{fig:asymmetryDisturbanceWeight}
        \end{figure*}

        The controller's robustness in executing multiple brachiation maneuvers is investigated through several tests. 
        Firstly, it is tested with the addition of extra weight, up to \SI{200}{\gram} is loaded on the tail (symmetric about the sagittal plane). 
        Secondly, its performance is evaluated when \SI{170}{\gram} is loaded on one of the arms (asymmetric about the sagittal plane). 
        Finally, the robustness is examined in the presence of a \SI{520}{\gram} disturbance placed in the robot path. 
        In the scenarios mentioned above, the controller effectively stabilized the trajectories, enabling RicMonk to perform multiple brachiation maneuvers in two out of three trials. 
        Fig. \ref{fig:asymmetryDisturbanceWeight} illustrates the robot performing three forward brachiation maneuvers as it carries \SI{80}{\gram} on one of the arms, i.e. asymmetric loading, and \SI{100}{\gram} on the tail. It also faces significant disturbances. 

    \subsection{Comparitive Analysis}
        AcroMonk is a minimalist brachiation robot, lightweight (\SI{1.6}{\kg}) and highly portable. 
        Its underactuated design employs a single actuator, enabling robust forward brachiation, but it struggles with backward motion and limited momentum due to its lightweight build.
        RicMonk, a three-link brachiation robot with a body-like structure, weighs \SI{3.3}{\kg}, offering portability and agility. 
        Featuring two actuators, it achieves versatile forward and backward brachiation, benefitting from its tail structure for enhanced energy input.
        A comparative analysis in terms of standard deviations for FR ($\boldsymbol{\sigma}^{\text{FR}}_{0}$) and BR ($\boldsymbol{\sigma}^{\text{BR}}_{0}$) between AcroMonk and RicMonk, indicates larger values for RicMonk. 
        However, it's crucial to approach these values objectively, which stems from several factors. 
        Firstly, RicMonk weighs twice as much as AcroMonk, and the torque used for front and back release in RicMonk is nearly three times greater than that in AcroMonk.
        Moreover, RicMonk possesses the capability to independently apply torque to each arm, a feature that is missing in AcroMonk. 
        Cost of Transport (CoT) \cite{costOfTransportWeb}, \cite{kim2017design} is a dimensionless measure for energy efficiency, allowing comparison across sizes and structures. 
        Formulated as Eq. \ref{cot}, energy input $E$, mass $m$, distance traveled $d$, and acceleration due to gravity $g$.      
        \begin{equation}
            CoT = \dfrac{E}{mgd}
            \label{cot}
        \end{equation}
        A lower CoT value indicates greater energy efficiency within a given system. 
        Table (\ref{table:energyCompare}) compares the total energy (TE) consumed by the AcroMonk and RicMonk as they perform five consecutive forward brachiation (BF) maneuvers (in joules), their CoT (dimensionless), and the time of transport ($t$) for the five maneuvers in seconds. 
        \begin{table}[H]
            \centering
            \caption{Comparitive analysis}
            \label{table:energyCompare}
            \begin{tabular}{lccc} 
                \toprule
                    & TE (J) & CoT & $t$ (s)\\ 
                \midrule
                AcroMonk & 8.9547 & 0.3355 & 11 \\ 
                RicMonk & 15.1947 & 0.2760 & 17 \\ 
                \bottomrule
            \end{tabular}            
        \end{table}
        
        The table highlights several key insights. 
        Though AcroMonk consumes a lot less energy in total, RicMonk has a lower CoT, indicating higher energy efficiency.
        However, these findings are limited to the trajectories that have been optimized and stabilized, and other trajectories might have different characteristics. 
        Also, CoT is one of many perspectives to compare energy efficiency among systems.

\section{\textsc{Conclusion and Outlook}}
\label{section:conclusion}
This paper addressed the mechatronic design, generation, and stabilization of optimal trajectories, and experimental validation of RicMonk, a three-link underactuated brachiation robot, with passive grippers. 
RicMonk is portable, modular, and easily reproducible. 
The presence of a tail structure and two actuators help RicMonk to perform multiple brachiation maneuvers in both forward and backward directions and this is a novelty in literature. 
This also presents a notable advantage when compared to its predecessor, AcroMonk. 
The presence of the tail also improves the energy efficiency as shown by the comparison of cost of transport. 
We are working on the use of a floating-base model to optimize trajectories for ricochetal brachiation. 
Additionally, brachiation over irregularly placed bars and online trajectory optimization are challenges that lie ahead. Further, the gripper design may be improved to require a lower amount of torque to perform release action. Enabling RicMonk to perform ricochetal brachiation is a very exciting outlook.



\bibliographystyle{IEEEtran}
\bibliography{references}

\end{document}